\documentclass{article}
\usepackage{spconf,amsmath,graphicx, multirow, multicol, booktabs}
\usepackage[numbers,sort&compress]{natbib}

\title{ADAPTIVE QUANTIZATION WITH MIXED-PRECISION BASED ON LOW-COST PROXY}
\name{Junzhe Chen,  Qiao Yang, Senmao Tian, Shunli Zhang$^{\ast}$ \thanks{*Corresponding author}}
\address{Beijing Jiaotong University\\
        School of Software Engineering\\
        Beijing, China}
\begin{document}
%
\maketitle
\begin{abstract}
It is critical to deploy complicated neural network models on hardware with limited resources. This paper proposes a novel model quantization method, named the Low-Cost Proxy-Based Adaptive Mixed-Precision Model Quantization (LCPAQ), which contains three key modules. 
The hardware-aware module is designed by considering the hardware limitations, while an adaptive mixed-precision quantization module is developed to evaluate the quantization sensitivity by using the Hessian matrix and Pareto frontier techniques. Integer linear programming is used to fine-tune the quantization across different layers. Then the low-cost proxy neural architecture search module efficiently explores the ideal quantization hyperparameters. 
Experiments on the ImageNet demonstrate that the proposed LCPAQ achieves comparable or superior quantization accuracy to existing mixed-precision models. Notably,  LCPAQ achieves 1/200 of the search time compared with existing methods, which provides a shortcut in practical quantization use  for resource-limited devices.
\end{abstract}
\section{Introduction}
\label{sec:intro}
The large scale and complexity of deep learning models often hinder their deployment, especially when considering advanced models with high computational demands. A promising approach to address this is through neural network quantization\cite{morgan, nagel2021white, krishnamoorthi2018quantizing, lsq+, chin, hawq, jacob2018quantization, kim2021zero, song2020drq, park2018energy, Tian2023CABMCB}, which seeks to minimize computational time and energy cost while preserving network precision. Mixed-precision quantization (MPQ) is presented to optimize the scale of the deep learning model, the computational demand, and the accuracy by assigning different bit widths to the layers. 

Recent studies have presented methods such as HAQ\cite{wang2019haq} and DNAS\cite{wu2018mixed} to address the MPQ challenge. However, the expansive search space makes it infeasible to evaluate every mixed-precision combination for optimal performance. HAWQv2\cite{dong2020hawq} determines quantization sensitivity by assessing the mean of the eigenvalues of the Hessian matrix. It then applies a Pareto frontier strategy for bit-width decision-making to avoid manual selection. But given the immense search space, HAWQv2 may not always identify the optimal configuration. To improve this, we integrate hardware-aware constraints into our optimal bit-width search and explore the impact of various quantization hyperparameters.

Hardware-aware ensures the model is optimized for specific hardware, boosting the performance. For example, HAWQv3\cite{yao2021hawq} provides an Integer Linear Programming (ILP) solution to generate mixed-precision configurations under various constraints like model size, bit operations (Bops), and latency. We use ILP as our hardware constraint for MPQ, integrating it to sensitivity measurements to refine the bit-width selection.

Our proposed solution, LCPAQ, combines a hardware-aware module, an adaptive mixed-precision quantization, and a low-cost proxy neural architecture search module. Our main contributions include:

(1) We present a method computing the Hessian matrix's trace to determine layer sensitivity and employ a Pareto frontier strategy for adaptive bit-width selection in mixed-precision quantization.

(2) Our model ensembles a hardware-aware module and, in conjunction with Integer Linear Programming (ILP), fine-tunes precision settings based on hardware limitations.

(3) We involve a low-cost proxy for Neural Architecture Search (NAS) to automate hyperparameter selection in model quantization, which accelerates the searching process.

(4) Thorough evaluations on the ImageNet dataset display that our LCPAQ model can achieve consistently comparable or superior results to advanced quantization methods. Subsequent ablation studies further highlight the efficiency of our approach.

\begin{figure*}[ht]
\vspace{-2.0em} 
\includegraphics[width=\linewidth, height=0.4\linewidth]{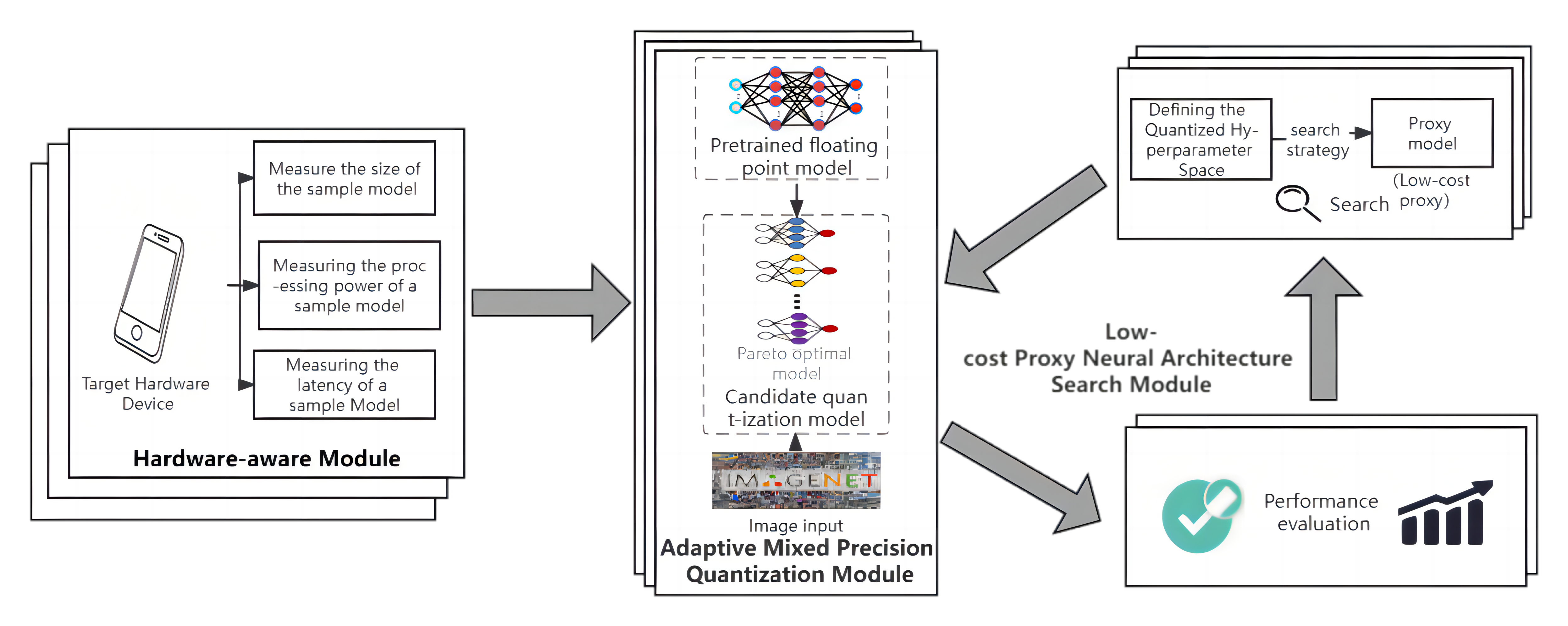}
\caption{Overall framework of LCPAQ model, including Hardware-aware Module, Adaptive Mixed Precision Quantization Module and Low-Cost Proxy Neural Architecture Search Module} \label{fig2}
\vspace{-1.0em} 
\end{figure*}

\section{LCPAQ MODEL}
\label{sec:format}

As shown in Figure \ref {fig2}, our model introduces an adaptive mixed-precision quantization approach based on low-cost proxy. Under hardware-aware conditions, our model trains efficient low-cost agents, searches for quantization-related hyperparameters and distillation, and employs automatic bit-width selection techniques.

\subsection{Hardware-aware Module}
\label{ssec:subhead}
In LCPAQ's hardware-aware module, three key metrics, i.e. model size, Bops, and latency, are evaluated for the target hardware. Direct hardware-level latency measurements are typically more accurate than computational estimates. When considering quantization to low precision, the trade-offs between accuracy and latency should be taken into consideration.

To determine the best bit-width precision for each layer, we construct the automatic bit-width selection within hardware constraints as an Integer Linear Programming (ILP) problem. For a network model with $ L $ layers and $ Z $ available precision for each layer's quantization, the ILP search space is $ Z^{L} $. The goal of the ILP is to find the best bit-width setting from these $ Z^{L} $  options, balancing model perturbations with user-defined constraints. Each precision setting might lead to different model perturbations. For simplicity's sake, we assume that perturbations across different layers are mutually independent, i.e., $\delta=\sum_{i=1}^{L}\delta_{i}^{(b_{i})}$, where $ \delta_{i} $ represents the perturbation of the $ i^{th} $ layer for Z  bit-width. This implies that one can measure the sensitivity for each layer independently, necessitating only $ L\times Z $ computations. Concerning sensitivity metrics, our work harnesses perturbations upon on the Hessian matrix\cite{dong2020hawq}. The ILP problem tries to find the right bit precision that minimizes this perturbation, as follows:
\setlength{\abovedisplayskip}{3pt}
\setlength{\belowdisplayskip}{3pt}
\begin{small} 
\begin{align*} 
Target: \; min_{\{b_{i}\}_{i=1}^{L}}&\sum_{i=1}^{L}\delta_{i}^{(b_{i})} \label{target} \tag{1} ,\\
Subject \; to: \sum_{i=1}^{L}S_{i}^{(b_{i})} &\leq Model \; Size \; Limit \tag{2} ,\\ 
\sum_{i=1}^{L}P_{i}^{(b_{i})} &\leq BOPS \; Limit \tag{3} ,\\
\sum_{i=1}^{L}T_{i}^{(b_{i})} &\leq Latency \; Size \; Limit. \tag{4}
\end{align*}
\end{small}
Where $ S_{i}^{(b_{i})} $  denotes the size of the $ i^{th} $ layer of the model quantized with Z bits; $ T_{i}^{(b_{i})} $ is the relative latency time; and $ p_{i}^{(b_{i})} $\cite{van2020bayesian} is employed to determine the requisite Bops for that layer, essentially measuring the total bit operations needed for processing a layer.

\subsection{Adaptive Mixed-Precision Quantization Module}
\label{ssec:subhead}

We determine the relative sensitivity rank of the model's layers by calculating the average trace of the Hessian matrix. Due to challenges in accessing the Hessian matrix directly, we use the randomized algorithm \cite{avron2011randomized, bai1996some}. This matrix-free technique eliminates the step to interact directly with the Hessian matrix, ensuring efficient calculations.

Here, we examine the trace of the symmetric matrix $ H\in R^{d \times d} $. Subsequently, considering a random vector $ z\in R^{d} $ that complies with an independent and identically distributed Gaussian distribution, N(0,1) (or Rademacher distribution), it can be formally expressed as:
\setlength{\abovedisplayskip}{3pt}
\setlength{\belowdisplayskip}{3pt}
\begin{small}
\begin{equation}
 \begin{split}
 Tr(H) &= Tr(HI) = Tr(HE \begin{bmatrix} zz^T \end{bmatrix}) \\
   &= E(Tr \begin{bmatrix} Hzz^T \end{bmatrix}) = E(\begin{bmatrix} z^T Hz \end{bmatrix})
 \end{split}
 \tag{5}
\end{equation}
\end{small}
where $ I $ denotes the identity matrix.

To achieve an accurate bit-precision setting, we assume that the different progressive quantization aware fine-tuning stages are represented by $ C $. After every $ C $ fine-tuning steps, the bit-width of the quantized model is reduced once. Finally,  the model is converged to the final precision. The entire search space can be written as $ Z \times C $:
\setlength{\abovedisplayskip}{3pt}
\setlength{\belowdisplayskip}{3pt}
\begin{small}
\begin{align}
Z = &m^{L} \tag{6} \\
C = \sum _{i=1}^{L}i! \times S&(L,i) \to L! \tag{7}    
\end{align}
\end{small}
Where $m$ denotes the number of the quantization precision options, $L$ is the number of layers in the given model, and $S$ represents the second kind of Stirling numbers.

We first use ILP to minimize model disturbance, considering other goals as constraints to form a baseline solution. From this starting point, we adjust the bit-width for a Pareto analysis to find a trade-off between multiple objectives. We usually adopt the solution with the highest precision within our constraints. Uniquely, our method combines ILP with the Pareto frontier, both reducing the computational demands of traditional ILP and focusing on the optimal solutions within the frontier.

\subsection{Low-Cost Proxy Neural Architecture Search Module}
\label{ssec:subhead}

The primary role of the low-cost proxy neural architecture search module is to help the model efficiently determine the best quantization hyperparameter configuration, minimizing computational and search expenses. The selection of these hyperparameters significantly impacts the model's performance and compression outcomes.

We use early stopping during training as our training strategy. Being hardware-aware, we define the hyperparameter search space and choose a search approach (e.g., evolutionary algorithms, reinforcement learning). Then we can generate and assess potential quantization architectures. The performance metrics computed on a validation set as feedbacks will further refine our search.

We start with M possible hyperparameter combinations. From these, N sets (where N $\ll$ M) are randomly chosen for initial short-term training. The configurations of these N models are encoded, serving as proxy model input. Upon establishing an initial proxy model, we utilize a search strategy to acquire new hyperparameter sets. These are then assessed by using the pre-trained proxy model. The top-performing K models (with K $<$ N) are chosen and trained to full capacity to determine their accuracy. Their results, combining with the original N models, refine the proxy model. Over time, as the proxy model is fed by more data, its accuracy improves. After a sequential exploration constrained by the size of the hyperparameter search space, we use the low-cost proxy to simulate the relative accuracy of the quantized model. Then we pick the best model, and fully retrain it to obtain the final model.

\subsection{Distillation in Quantization}
\label{ssec:subhead}

Previous work\cite{polino2018model} emphasized that knowledge distillation can make up the performance decline in low-bit quantization (like 4-bit, 2-bit, or mixed-bit). Hence, in shaping our hyperparameter search space, we integrated the distillation approach. Our results are similar to previous studies, confirming that knowledge distillation increases quantized model accuracy. For our ImageNet image classification experiments, we adopt the straightforward distillation method\cite{hinton2015distilling} to simplify the process.

\section{EXPERIMENTS}
\label{sec:result}

All experiments were standardized with a learning rate of $ 1 \times 10^{-4} $, a weight decay coefficient of $ 1 \times 10^{-4} $, a batch size of 128. The optimizer is set to SGD.

For the NAS, a simple random search strategy is used in the search phase. The initial training phase consisted of 10 epochs, followed by 100 epochs in subsequent phases. We utilized a straightforward two-layer MLP. The topK parameter for proxy model selection was set at 3.

\begin{table}[h]
\vspace{-1.0em}
  \begin{center}
    \caption{Mixed precision with different constraints for ResNet18 hardware-aware}
    \renewcommand{\arraystretch}{1}
    \label{tab1}
    \vspace{0.5em}
    \begin{tabular}{c c c c c} 
    \toprule[1.5pt]
      \textbf{} & \textbf{Level} & \textbf{Size} & \textbf{Bops} & \textbf{Accuracy}\\
      \midrule
      INT8 & - & 11.2 & 114 & 72.64 \\
      \midrule
      \multirow{3}{*}{Model-size} & High & 9.9 & 103 & 72.07\\ %
      & Medium & 7.9 & 98 & 71.38 \\ %
      & Low & 7.3 & 95 & 70.67\\
      \midrule
      \multirow{3}{*}{Bops} & High & 8.7 & 92 & 71.36\\ %
      & Medium & 6.7 & 72 & 68.86 \\ %
      & Low & 6.1 & 54 & 65.09 \\
      \midrule
      \multirow{3}{*}{Latency} & High & 8.7 & 92 & 65.08\\ %
      & Medium & 7.2 & 76 & 69.87 \\ %
      & Low & 6.1 & 54 & 65.54 \\
      \midrule
      INT4 & - & 5.6 & 28 & 60.21 \\
    \bottomrule[1.5pt]
    \end{tabular}
  \end{center}
\vspace{-2.0em}  
\end{table}

\begin{table}[h]
\vspace{-1.0em}
  \begin{center}
    \caption{Mixed precision with different constraints for ResNet50 hardware-aware}
    \renewcommand{\arraystretch}{1}
    \label{tab2}
    \vspace{0.5em}
    \begin{tabular}{c c c c c} 
    \toprule[1.5pt]
      \textbf{} & \textbf{Level} & \textbf{Size} & \textbf{Bops} & \textbf{Accuracy}\\
      \midrule
      INT8 & - & 24.5 & 247 & 77.07 \\
      \midrule
      \multirow{3}{*}{Model-size} & High & 21.3 & 226 & 76.61\\ %
      & Medium & 19.0 & 197 & 76.07 \\ %
      & Low & 16.0 & 168 & 74.58\\
      \midrule
      \multirow{3}{*}{Bops} & High & 22.0 & 197 & 76.42\\ %
      & Medium & 18.7 & 154 & 75.51 \\ %
      & Low & 16.7 & 110 & 73.21 \\
      \midrule
      \multirow{3}{*}{Latency} & High & 22.3 & 199 & 76.51\\ %
      & Medium & 18.5 & 155 & 75.30 \\ %
      & Low & 16.5 & 114 & 73.22 \\
      \midrule
      INT4 & - & 13.1 & 67 & 63.34 \\
    \bottomrule[1.5pt]
    \end{tabular}
  \end{center}
\vspace{-2.0em}  
\end{table}

\subsection{Mixed-precision Results with Different Constraints}
\label{ssec:subhead}

We propose the integration of hardware-aware constraints during quantization. The ILP problem outlined in equation \ref{target} highlights three pivotal constraints: model size, Bops, and latency. Our experiment set three different thresholds for each constraint, aiming to understand how ILP balances them to optimize the quantized model, thus improving its accuracy and performance. The adaptability of the ILP formula is evident in its ability to focus on a single constraint or a mix.

Tables \ref{tab1} to \ref{tab2} present outcomes for individual constraints and are divided into three sections: model size, Bops, and latency. Each section reflects the constraints set by experts, using varied restriction levels to represent different hardware limitations. Drawing insights from the data presented in Tables 1 and 2, we can draw the following conclusions:

(1) Model size doesn't strongly correlate with Bops. For example, the 'Medium' Model-size tier of ResNet18, despite being larger, has lower Bops than its 'High' Bops counterpart.

(2) Model size isn't directly linked to quantization precision. For ResNet50, a 'Low' Bops model (16.7MB) achieves 73.21$\%$ accuracy, whereas a 'Low' Model-size version (16.0MB) has a better accuracy of 74.58$\%$.

(3) The floating-point precision benchmark for ResNet18 is 71.47 with a size of 44.6MB. Comparatively, ResNet50 achieves greater precision with a smaller size across all quantization. This suggests, that for hardware-limited scenarios, it's wiser to quantize complex models over simpler ones.

(4) Transitioning from ResNet18 to ResNet50 increases model complexity but improves the generalization ability of the model. This shields the model during low-precision quantization from the negative impacts of poorly chosen bit-widths in sensitive layers.

\subsection{Mixed-precision quantization results under different quantization hyperparameter settings}
\label{ssec:subhead}

In our study, we examine the effects of various quantization hyperparameters on model precision. We investigate five conditions: Uniform Quantization at both 8-bit and 4-bit, constraints set to Medium at 0.5 for Model-Size, Bops, and Latency. Yet, only the results for 8-bit Uniform Quantization are presented.

Given the expansive search space, it's impractical to detail the precision of every quantized model on ImageNet classification tasks. Instead, Table 3 focuses on three key quantization hyperparameters: Channel: 'Y' for per-channel quantization and 'N' for per-tensor. BN: 'Y' indicates the use of Batch Normalization (BN) layer folding, while 'N' denotes its absence. Distil: 'Y' for employing knowledge distillation techniques, 'N' for not. 

Although other hyperparameters, like clipping range quantization types and unique formats for weights and activations, were part of our study, their specific outcomes aren't detailed here for brevity. From Table \ref{tab3}, the following observations can be inferred:

\begin{table}[h]
\vspace{-1.0em}
  \begin{center}
    \caption{Mixed precision of ResNet18 under different quantization hyperparameter settings under Int8}
    \renewcommand{\arraystretch}{1}
    \label{tab3}
    \vspace{0.5em}
    \begin{tabular}{c c c c c} 
    \toprule[1.5pt]
      \textbf{ResNet18} & \textbf{Channel} & \textbf{BN} & \textbf{Distill} & \textbf{Accuracy}\\
      \midrule
      \multirow{8}{*}{INT8} & Y & Y & Y & 72.81\\ %
      & Y & Y & N & 72.64 \\ %
      & Y & N & Y & \pmb{73.04}\\ %
      & Y & N & N & 73.02 \\ %
      & N & Y & Y & 72.74 \\ %
      & N & Y & N & 72.62 \\ %
      & N & N & Y & 72.77 \\ %
      & N & N & N & 72.49 \\ %

    \bottomrule[1.5pt]
    \end{tabular}
  \end{center}
\vspace{-1.0em}  
\end{table}

(1) Introducing knowledge distillation during the INT8 quantization process can improve the accuracy of the model to a certain extent.

(2) For a higher-precision quantization such as INT8 quantization, the selection of different quantization hyperparameters has little impact on the accuracy of the experimental results, because the accuracy of INT8 quantization itself will not be much difference compared to the floating point model accuracy benchmark.

(3) Using BN layer folding technology in quantization usually results in some loss of precision, but it also speeds up model training and inference stages.  It's worth noting that the sensitivity towards BN layer folding may vary across different network architectures and tasks.

Optimal quantization hyperparameter setting can effectively reduce the negative impact of poorly set sensitive layers. Using low-cost proxy models significantly boosts the efficiency of the quantization process. This method not only saves computational resources but also time. According to preliminary statistics,even with a small search space defined by quantitative hyperparameters, using a proxy model takes only 1/200th of the time compared to manual setups. As the search space further expands, this time advantage will become more obvious. This high efficiency is mainly reflected in the significant reduction of experiment time and computing resources for large-scale search.

\section{CONCLUSIONS}
\label{sec:conlution}

We propose a low-cost proxy-based approach for adaptive mixed-precision model quantization named LCPAQ. By merging adaptive quantization techniques with low-cost proxy models, our research improves quantization efficiency. Experiments on the ImageNet dataset demonstrate the LCPAQ model can improve quantization accuracy and simplify the search for optimal quantization parameters. Notably, our method achieves comparable performance to existing quantization methods while providing quicker and more intuitive parameter configuration. 

However, LCPAQ is designed mainly for image classification tasks. Future adaptations could encompass tasks like segmentation, object detection, or natural language processing. Furthermore, the proxy model might not consistently reflect the actual model's performance. These challenges give directions for future exploration of our proposed method.

\vfill\pagebreak

\bibliographystyle{IEEEbib}
\bibliography{strings,refs}

\begin{thebibliography}{10}

\bibitem{morgan}
Nelson Morgan et~al.,
\newblock ``Experimental determination of precision requirements for
  back-propagation training of artificial neural networks,''
\newblock in {\em Proc. Second Int’l. Conf. Microelectronics for Neural
  Networks}. Citeseer, 1991, pp. 9--16.

\bibitem{nagel2021white}
Markus Nagel, Marios Fournarakis, Rana~Ali Amjad, Yelysei Bondarenko, Mart
  Van~Baalen, and Tijmen Blankevoort,
\newblock ``A white paper on neural network quantization,''
\newblock {\em arXiv preprint arXiv:2106.08295}, 2021.

\bibitem{krishnamoorthi2018quantizing}
Raghuraman Krishnamoorthi,
\newblock ``Quantizing deep convolutional networks for efficient inference: A
  whitepaper,''
\newblock {\em arXiv preprint arXiv:1806.08342}, 2018.

\bibitem{lsq+}
Yash Bhalgat, Jinwon Lee, Markus Nagel, Tijmen Blankevoort, and Nojun Kwak,
\newblock ``Lsq+: Improving low-bit quantization through learnable offsets and
  better initialization,''
\newblock in {\em Proceedings of the IEEE/CVF Conference on Computer Vision and
  Pattern Recognition Workshops}, 2020, pp. 696--697.

\bibitem{chin}
Ting-Wu Chin, Pierce I-Jen Chuang, Vikas Chandra, and Diana Marculescu,
\newblock ``One weight bitwidth to rule them all,''
\newblock in {\em Computer Vision--ECCV 2020 Workshops: Glasgow, UK, August
  23--28, 2020, Proceedings, Part V 16}. Springer, 2020, pp. 85--103.

\bibitem{hawq}
Zhen Dong, Zhewei Yao, Amir Gholami, Michael~W Mahoney, and Kurt Keutzer,
\newblock ``Hawq: Hessian aware quantization of neural networks with
  mixed-precision,''
\newblock in {\em Proceedings of the IEEE/CVF International Conference on
  Computer Vision}, 2019, pp. 293--302.

\bibitem{jacob2018quantization}
Benoit Jacob, Skirmantas Kligys, Bo~Chen, Menglong Zhu, Matthew Tang, Andrew
  Howard, Hartwig Adam, and Dmitry Kalenichenko,
\newblock ``Quantization and training of neural networks for efficient
  integer-arithmetic-only inference,''
\newblock in {\em Proceedings of the IEEE conference on computer vision and
  pattern recognition}, 2018, pp. 2704--2713.

\bibitem{kim2021zero}
Sungrae Kim and Hyun Kim,
\newblock ``Zero-centered fixed-point quantization with iterative retraining
  for deep convolutional neural network-based object detectors,''
\newblock {\em IEEE Access}, vol. 9, pp. 20828--20839, 2021.

\bibitem{song2020drq}
Zhuoran Song, Bangqi Fu, Feiyang Wu, Zhaoming Jiang, Li~Jiang, Naifeng Jing,
  and Xiaoyao Liang,
\newblock ``Drq: dynamic region-based quantization for deep neural network
  acceleration,''
\newblock in {\em 2020 ACM/IEEE 47th Annual International Symposium on Computer
  Architecture (ISCA)}. IEEE, 2020, pp. 1010--1021.

\bibitem{park2018energy}
Eunhyeok Park, Dongyoung Kim, and Sungjoo Yoo,
\newblock ``Energy-efficient neural network accelerator based on outlier-aware
  low-precision computation,''
\newblock in {\em 2018 ACM/IEEE 45th Annual International Symposium on Computer
  Architecture (ISCA)}. IEEE, 2018, pp. 688--698.

\bibitem{Tian2023CABMCB}
Senmao Tian, Ming Lu, Jiaming Liu, Yandong Guo, Yurong Chen, and Shunli Zhang,
\newblock ``Cabm: Content-aware bit mapping for single image super-resolution
  network with large input,''
\newblock {\em 2023 IEEE/CVF Conference on Computer Vision and Pattern
  Recognition (CVPR)}, pp. 1756--1765, 2023.

\bibitem{wang2019haq}
Kuan Wang, Zhijian Liu, Yujun Lin, Ji~Lin, and Song Han,
\newblock ``Haq: Hardware-aware automated quantization with mixed precision,''
\newblock in {\em Proceedings of the IEEE/CVF conference on computer vision and
  pattern recognition}, 2019, pp. 8612--8620.

\bibitem{wu2018mixed}
Bichen Wu, Yanghan Wang, Peizhao Zhang, Yuandong Tian, Peter Vajda, and Kurt
  Keutzer,
\newblock ``Mixed precision quantization of convnets via differentiable neural
  architecture search,''
\newblock {\em arXiv preprint arXiv:1812.00090}, 2018.

\bibitem{dong2020hawq}
Zhen Dong, Zhewei Yao, Daiyaan Arfeen, Amir Gholami, Michael~W Mahoney, and
  Kurt Keutzer,
\newblock ``Hawq-v2: Hessian aware trace-weighted quantization of neural
  networks,''
\newblock {\em Advances in neural information processing systems}, vol. 33, pp.
  18518--18529, 2020.

\bibitem{yao2021hawq}
Zhewei Yao, Zhen Dong, Zhangcheng Zheng, Amir Gholami, Jiali Yu, Eric Tan,
  Leyuan Wang, Qijing Huang, Yida Wang, Michael Mahoney, et~al.,
\newblock ``Hawq-v3: Dyadic neural network quantization,''
\newblock in {\em International Conference on Machine Learning}. PMLR, 2021,
  pp. 11875--11886.

\bibitem{van2020bayesian}
Mart Van~Baalen, Christos Louizos, Markus Nagel, Rana~Ali Amjad, Ying Wang,
  Tijmen Blankevoort, and Max Welling,
\newblock ``Bayesian bits: Unifying quantization and pruning,''
\newblock {\em Advances in neural information processing systems}, vol. 33, pp.
  5741--5752, 2020.

\bibitem{avron2011randomized}
Haim Avron and Sivan Toledo,
\newblock ``Randomized algorithms for estimating the trace of an implicit
  symmetric positive semi-definite matrix,''
\newblock {\em Journal of the ACM (JACM)}, vol. 58, no. 2, pp. 1--34, 2011.

\bibitem{bai1996some}
Zhaojun Bai, Gark Fahey, and Gene Golub,
\newblock ``Some large-scale matrix computation problems,''
\newblock {\em Journal of Computational and Applied Mathematics}, vol. 74, no.
  1-2, pp. 71--89, 1996.

\bibitem{polino2018model}
Antonio Polino, Razvan Pascanu, and Dan Alistarh,
\newblock ``Model compression via distillation and quantization,''
\newblock {\em arXiv preprint arXiv:1802.05668}, 2018.

\bibitem{hinton2015distilling}
Geoffrey Hinton, Oriol Vinyals, and Jeff Dean,
\newblock ``Distilling the knowledge in a neural network,''
\newblock {\em arXiv preprint arXiv:1503.02531}, 2015.

\end{thebibliography}

\end{document}